\documentclass[runningheads]{llncs}

\usepackage{eccv}

\usepackage{eccvabbrv}

\usepackage{multirow}
\usepackage{multicol}
\usepackage{xcolor}
\usepackage{colortbl}
\usepackage{marvosym}

\usepackage{graphicx}
\usepackage{booktabs}

\usepackage[accsupp]{axessibility}  

\usepackage{hyperref}

\usepackage{orcidlink}

\makeatletter
\def\thanks#1{\protected@xdef\@thanks{\@thanks
        \protect\footnotetext{#1}}}
\makeatother

\begin{document}

\title{Just a Hint: Point-Supervised Camouflaged Object Detection} 

\titlerunning{Just a Hint: Point-Supervised Camouflaged Object Detection}

\author{Huafeng Chen\inst{1,2} \and Dian Shao\inst{1,2} \textsuperscript{\Letter}\thanks{\Letter \ Corresponding Authors} \and Guangqian Guo\inst{1,2} \and Shan Gao\inst{1,2} \textsuperscript{\Letter}}

\authorrunning{H.~Chen et al.}

\institute{$^{1}\ $Unmanned System Research Institute, Northwestern Polytechnical University\\ $^{2}\ $National Key Laboratory of Unmanned Aerial Vehicle Technology
\\ \email{\{chf,guogq21\}@mail.nwpu.edu.cn, \{shaodian,gaoshan\}@nwpu.edu.cn} }

\maketitle

\begin{abstract}
  Camouflaged Object Detection (COD) demands models to expeditiously and accurately distinguish objects which conceal themselves seamlessly in the environment. Owing to the subtle differences and ambiguous boundaries, COD is not only a remarkably challenging task for models but also for human annotators, requiring huge efforts to provide pixel-wise annotations. To alleviate the heavy annotation burden, we propose to fulfill this task with the help of only one point supervision. Specifically, by swiftly clicking on each object, we first adaptively expand the original point-based annotation to a reasonable hint area. Then, to avoid partial localization around discriminative parts, we propose an attention regulator to scatter model attention to the whole object through partially masking labeled regions. Moreover, to solve the unstable feature representation of camouflaged objects under only point-based annotation, we perform unsupervised contrastive learning based on differently augmented image pairs (e.g. changing color or doing translation). On three mainstream COD benchmarks, experimental results show that our model outperforms several weakly-supervised methods by a large margin across various metrics. 
\end{abstract}

\section{Introduction}

\begin{figure}[t]
\centering
\includegraphics[width=12cm]{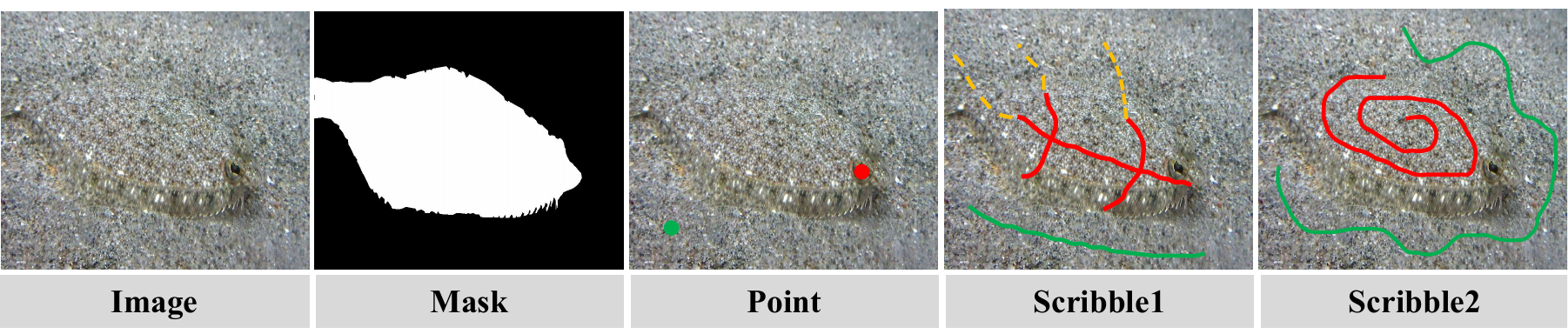}\\
\caption{Different types of annotation in camouflaged object detection task. The mask annotation takes about 60 minutes for each image. The scribbles take about 10 seconds but have diversity and boundary issues. The point takes just 2 seconds and only needs to point the most discriminative part of the camouflaged object.}

\label{fig:1}
\end{figure}

Camouflaged Object Detection (COD) aims to distinguish the disguised objects hiding carefully in the environment~\cite{pang2022zoom,fan2020camouflaged,fan2021concealed,fan2023advances}. Recently, it has attracted ever-growing research interest from the computer vision community and facilitates valuable real-life applications, such as search and rescue~\cite{fan2021concealed}, species discovery~\cite{perez2012early}, and medical image analysis~\cite{fan2020inf, fan2020pranet}.
Specifically, 
we tentatively figure out \textit{two key-points} for this task:
1) \textit{Quickly probing} the existence of camouflaged objects, and 
2) \textit{Accurately detecting} the corresponding locations pixel-wise within the given image.
Towards this end,
accurate pixel-level annotations are required,
based on which several fully-supervised approaches achieve good performance \cite{lv2021simultaneously,fan2020camouflaged}.
However, 
since spotting camouflaged objects is also difficult for humans,
such an annotation process is not only laborious but also time-consuming, \emph{e.g.}, 60 minutes for each image~\cite{fan2020camouflaged}.

To address this issue,
weakly-supervised COD has been explored recently.
In~\cite{he2023weakly}, the authors try to use scribbles to replace pixel-wise annotations and propose a novel CRNet to solve this problem.
But such a setting falls into a dilemma:
1) efficiency, the quality of scribble could not be ensured, arbitrary patterns could harm the learning process, 
2) effectiveness, scribbles are expected to indicate the outline of boundaries, which also increases the annotation difficulty, as shown in Fig. \ref{fig:1}.



In this work, we propose to fulfill the COD task with only point-based supervision.
Such an idea is motivated by the cognitive process of the human vision system:
we tend to recognize the most “discriminative part” of one camouflaged object \cite{lv2021simultaneously}, and then scatter our attention to figure out the whole object mask.
Even sometimes the camouflaged object is not perceived at first glance,
we could quickly notice the target with \textit{just a little hint} when a friend points to the image and says “\textit{carefully scrutinize there!}”.
Importantly, the whole point-supervised annotation process only takes within $2$ seconds.



\begin{figure}[t]
\centering
\includegraphics[width=12cm]{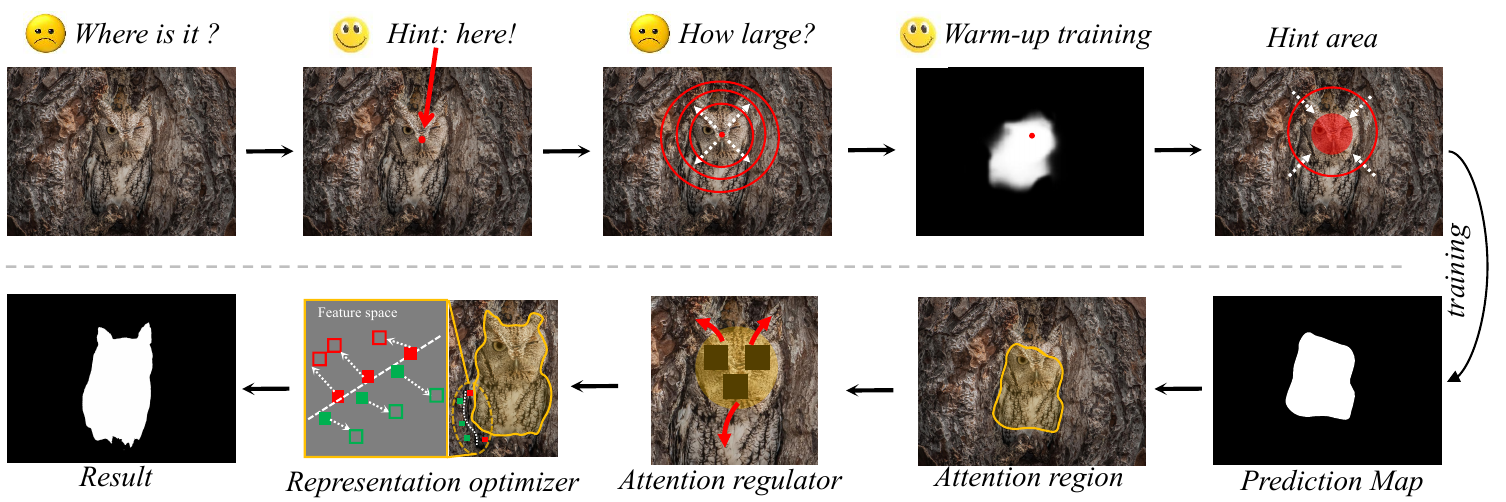}\\
\caption{Overview of our method. The first row shows the process of hint area generation from just a point to a hypothetical area. The second row shows the training process of attention regulator and presentation optimizer from the hint area to an accurate mask.
}
\label{fig:2-2}
\end{figure}

However, supervision provided by a single point is rather limited,
making the model easily collapse during training.
In a similar but different task, Salient Object Detection (SOD),
widely adopted solutions include detecting the edge map~\cite{gao2022weakly} first,
or generating the prediction map~\cite{wei2021shallow} as the proposal region at the first stage.
Unfortunately, 
all the previously mentioned solutions for generating proposal regions in SOD are not applicable in point-supervised COD,
owing to the low contrast and ambiguous edges between the camouflaged objects and backgrounds (shown in S.M.).
To address it,
we explore a new point-to-region supervision generation strategy, hint area generator.
Specifically,
starting from the single point annotation,
we adaptively expand it to a certain region, which obviously enriches the limited supervision, as shown in Fig. \ref{fig:2-2}.

Moreover, weakly-supervised COD methods suffer from partial detection, i.e., only discriminative parts of the object could gain adequate attention, 
while the whole shape is not accurately perceived, especially with the supervision of a single point.
To make the model not only focus on local distinct ones, but nondiscriminatory parts, we propose an attention regulator module.
By masking labeled region randomly, the model could scatter attention to the whole object, as shown in Fig. \ref{fig:2-2}.

\begin{figure}[t]
\centering
\includegraphics[width=12cm]{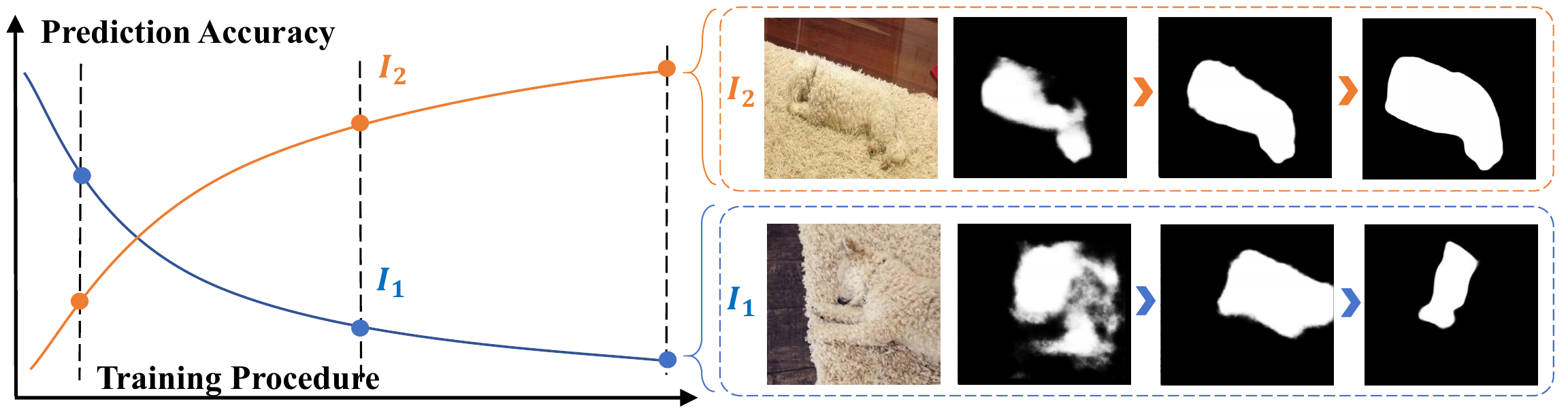}\\
\caption{Comparison of prediction accuracy for similar images during training. Although two images $I_1$ and $I_2$ are very similar, their prediction accuracy trends are opposite. It is largely due to 
that the learned features are not robust enough in weak supervision. 
}

\label{fig:2}
\end{figure}

Furthermore, the learned feature representations with only point-based annotation tend to be unstable. 
As shown in Fig.\ \ref{fig:2}, 
although the two images appear to be similar, the prediction accuracy largely varies. 
This is largely due to that the learned features are not robust enough.
To obtain invariant patterns and stable representations during training,
we apply Unsupervised Contrastive Learning (UCL) strategy on differently augmented data pairs. 
Such UCL process facilitates the model to learn more reliable feature representations under point supervision. 

To summarize, our contributions are as follows:

\begin{itemize}
\item {We propose a novel point-based learning paradigm for the challenging COD task, and construct the first weakly-supervised COD dataset with point annotation, P-COD, which contains 3040 images from the training set of COD10K and 1000 images from the training set of CAMO. Annotators are only required to swiftly click a point within the camouflaged object according to their first impression without localizing the whole object.}



\item {We develop a point-supervised COD method by imitating the cognitive process of the human vision system, which contains several novel designs, 1) hint area generator, 2) attention regulator, and 3) representation optimizer.}

\item {Experimental results show that our method outperforms existing weakly-supervised COD methods largely and even surpasses many fully-supervised methods. Moreover, when migrated to scribble-supervised COD and SOD tasks, our method also achieves competitive results.
}
\end{itemize}

\section{Related Work}
\textbf{Weakly Supervised Camouflaged Detection.} Pixel-level annotation in COD is very time-consuming, in order to reduce the cost of labeling, 
recent work has begun to explore weakly supervised camouflaged object detection. CRNet\ \cite{he2023weakly} firstly tries to use scribble labels to train the model. However, scribble labels have the problems of label diversity and poor control of label quality. We try to use simpler and more natural point labels.

\noindent\textbf{Point Annotation.} There have been several works on point annotations in weakly-supervised segmentation\ \cite{bearman2016s,qian2019weakly,cheng2022pointly,shin2021all} and instance segmentation\ \cite{liew2017regional,benenson2019large}. Semantic segmentation focuses on class-specific categories, which is different from COD task. PSOD\ \cite{gao2022weakly} proposes a novel weakly-supervised Salient Object Detection (SOD) method using point supervision. 
But the drawbacks of it are also obvious: 1) extra edge detectors and edge information 2) flood fill algorithm introducing additional computation, and not appropriate for COD since camouflaged objects are not salient.


\noindent\textbf{Contrastive Learning.} The core of contrastive learning is to attract the positive sample pairs and repulse the negative sample pairs. Moco\ \cite{he2020momentum} maintained a queue of negative samples and turned one branch of siamese network into a momentum encoder to improve the consistency of the queue. BYOL \cite{grill2020bootstrap} proposed a self-supervised image representation learning approach, achieving state-of-the-art results without negative samples. The mainstream paradigm of contrastive learning later shifted to only using positive samples. SimSiam\ \cite{chen2021exploring} built an efficient and simple contrastive learning framework only using positive samples, which is free from the limitations of large batch size, momentum encoder, and negative samples. The above work \cite{he2020momentum,grill2020bootstrap,chen2021exploring} has proved that UCL can completely approach or even exceed the effect of supervised manner, and learn representations with strong generalization. Therefore, we try to introduce unsupervised contrastive learning to point-supervised COD to strengthen the representation of the model.

\begin{figure*}[t]
\centering
\includegraphics[width=12cm]{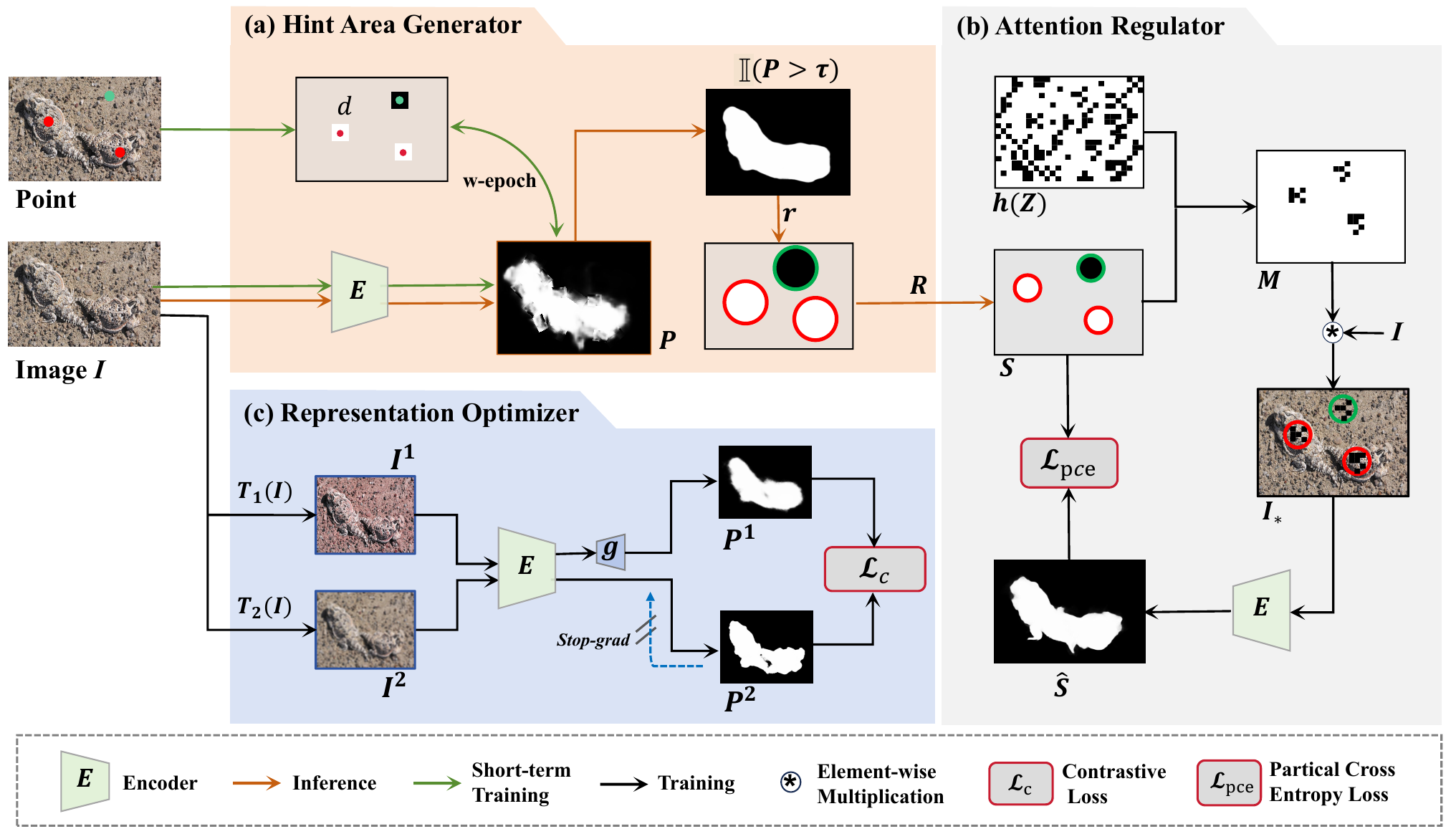}\\
\caption{Details of our method. The main components consist of three parts: (a) hint area generator, (b) attention regulator, and (c) representation optimizer. Hint area generator extends the point label from a single point to the small hint region.
Attention regulator enforce the model to focus on the whole object instead of being stuck in most discriminative part.
Representation optimizer uses the unsupervised contrastive learning to learn a stable feature representation for disentangling camouflaged objects and background.}
\label{fig:4}
\end{figure*}

\section{Methodology}

The overall architecture of the proposed framework is shown in Fig.\ \ref{fig:4}, which includes 
a) a \textit{hint area generator} to extend supervision from a single point to the small  hint region,
b) an \textit{attention regulator} to guide the model to focus on the whole object instead of being stuck in local discriminative parts, and
c) a \textit{representation optimizer} based on unsupervised contrastive learning for disentangling camouflaged objects and background.





%
\subsection{Hint Area Generator}

Compared with other forms of mask supervision, supervision provided by a point is rather limited,
which easily leads to model collapse, as shown in Tab.\ \ref{tab:3}.
To explain a little, an annotated point just functions like a \textit{“hint”},
only indicating the existence of the object,
with no knowledge of the area, shape, or boundary.
To solve this,
we design a hint area generator to adaptively expand a single point to a relatively larger region.
One of the most crucial things is \textit{to control the degree of point expansion}, avoiding the resultant hint area being too small (limited effect) or too large (out of boundary). Especially avoiding excessive expansion to introduce noise, place quality over quantity, because incorrect supervision is fatal for WSCOD, as shown in Tab.\ \ref{tab:3}.

Specifically, as shown in Fig.\ \ref{fig:4}(a), 
we initialize the desired annotation region with a small square area around the annotated point. It just prevents the model training from collapsing if only using one pixel as a point. 
The side length of the square area is $d$. 
Then, after training the encoder $w$ epochs as a warm-start (Short-term training), we hope this encoder could have a certain extent of semantic sense towards the object.
Based on this, we estimate the pseudo area size of the camouflaged object,
which is calculated by:
   ${P = E_{w}(I)}$,
where $E_{w}$ denotes the encoder after training $w$ epochs, $I$ represents the input image, and $P$ is the obtained prediction map used to serve as the rough object area hint.

We choose circular to be the basic area shape,
which could be intuitively regarded as \textit{``a large point"}.
Then the key is to determine the reasonable radius of the circular area.
Based on the object area hint $P$, the initialized radius $r$ is defined as:
\begin{equation}
   {r = \sqrt{\frac{Sum(\mathbb{I}(P>\tau))}{N}}},
\end{equation} 
where $\tau$ is a threshold, $\mathbb{I}(\cdot)$ is an indicator function, and $N$ is the number of objects. Note that $N$ could be achieved by counting the annotated point and thus is a deterministic parameter. 

Then, to prevent the calculated circular from being out of the object boundary,
we use a hyperparameter, $\alpha$, to alter the radius $R$ using:
   ${R = \frac{r}{\alpha}}$.
Specifically, the annotation for an object $n$ is enlarged from a single point $(x_n,y_n)$, to a small circle region around that point with the estimated radius $R$, denoted as $C^{R}_{(x_n,y_n)}$.
It is noteworthy that the background is also annotated by such a small circular region, unfolding around the point $(x_b,y_b)$, which is randomly picked from the background region.

As a result, with the help of the hint area generator, the original point supervision for the image $I$ is enriched into several small regions, expanding supervision while minimizing the introduction of incorrect supervision as much as possible, represented as:
\begin{equation}
   S(I)=C^{R}_{(x_b,y_b)} \cup \sum_{n} C^{R}_{(x_n,y_n)}, \quad n=1,...,N
\end{equation} 
where $S(I)$ is the obtained supervision, as shown in Fig. \ref{fig:4}.

\subsection{Attention Regulator}
Weakly-supervised methods often suffer from partial detection problems,
owing to the fact that model tends to focus too much on those discriminative regions. Inspired by the HaS method~\cite{kumar2017hide}, the random mask is able to inhibit the strong response, 
we further design an attention regulator module for PCOD. Because in camouflaged scenes, people often annotate point labels on the most visually discriminative regions for the camouflaged objects, we leverage this prior to guide mask generation to scatter the attention from the directly supervised discriminative part to the whole object area, as shown in Fig.~\ref{fig:4}(b).
Firstly, based on the given supervision region, $C^R$,
we could obtain a mask $M$:
\begin{equation}
M= \begin{cases}
M_i=h(Z),\quad &i\in C^R \\
M_i=1,\quad &i\notin C^R
\end{cases} 
\end{equation}
where $Z$ is a logical matrix consisting of zeros and ones with the same shape as $I$, $h$(·) denotes a shuffle operation, $i$ is the pixel index.

Then, we apply this mask to the original image $I$:
    $I_{*} = I \ast  M$,
where $\ast $ indicates element-wise multiplication. Through such formulation, the annotated 
discriminative area will be randomly masked. 
During the training process, the model needs to explore the recognition of the surrounding foreground areas of the mask to restore recognition of the masked foreground area, helping the model scatter attention to other regions within the foreground objects.
\subsection{Representation Optimizer}

In COD, the foreground and the background share high similarity.
The subtle difference sometimes comes from the texture abruption or the 
 color inconsistency.
Based on such observations, we hope at least from one aspect the discrepancy could be exposed.

Toward this goal, we propose to use unsupervised contrastive learning strategy to optimize the feature space,
aiming to make the foreground object distinguished from the background through the learned invariant patterns and stable representations. 


Specifically, as shown in Fig.\ \ref{fig:4}(c),
for image $I$, we adopt two different visual transformations $T_{1}$ and $T_{2}$, selecting from colorjitter, Gaussianblur, flip, etc. These visual transformations are able to change the appearance of images. For example, Gaussian blur will smooth the texture, and changing color is relatively harmless to the texture patterns.
This process results in two transformed images $ I^{1}$ and $ I^{2}$:
    $I^{1}=T_{1}(I)$, and $I^{2}=T_{2}(I)$.

We encode the transformed images differently into two prediction maps $P^{(1)}$ and $P^{(2)}$, denoting as
$P^{1}=g(f(I^{1}))$, and $P^{2}=f(I^{2})$,     
where $f$ represents the encoder to be learned, 
and $g$ is a small network cascaded over $f$ to increase encoding diversity.
When a position-related or size-related transformation $T$ (e.g., scale, crop) is applied to the image $I^1$, this $T$ should be applied to $P^2$ to be aligned with $P^1$, and vice versa.
Our objective is to minimize the distance between these two prediction maps:
\begin{equation} 
   \min \mathcal{D}(P^{1},P^{2})=\sum_{i}|P^{1}_i - P^{2}_i|,
\end{equation} 
where $i$ is the pixel index. Here we follow the design of UCL, i.e., stopping the gradient ($stopgrad$) update at one end, 
so the contrastive loss function is defined as:
\begin{equation} 
    \mathcal{L}_{c}=\mathcal{D}(P^{1},stopgrad(P^{2})). \label{eq:8}
\end{equation} 

The prediction gap using different transformations will be narrowed by minimizing the above loss,
making the learned feature representation robust.

\subsection{Network}
\textbf{Encoder.} The structure of our feature encoder is simple compared to previous work. Specifically, in order to capture long-distance feature dependency and obtain multi-scale information, we use PVT \cite{wang2021pyramid} as the backbone, for an input image $I\in \mathbb{R}^{3\times H\times W}$, we put it into the backbone to get the output features $F_{eat_{i}}$ for the $i$-th. Then, we get the multi-scale features ($F_{eat1}, F_{eat2}, F_{eat3}, F_{eat4}$) with ($\frac{1}{4}, \frac{1}{8}, \frac{1}{16}, \frac{1}{32}$) resolution of input image $I$. We downsize the channel dimension of $F_{eat_{i}}$ into $64$ by using $3\times3$ convolutional layers. Next, these feature maps are unified into the same size by an up-sampling operation, and combined through the concatenation. Finally the output map $\hat{S} {\in \mathbb{R}^{1\times W \times  H}}$ is obtained by the $3\times3$ convolution layer, and the detailed network structure is provided in supplementary materials.

\noindent\textbf{Loss.} Compared to other weak-supervised methods \cite{he2023weakly,yu2021structure,zhang2020weakly,wang2023temporal}, we have fewer losses, only two. The final loss $\mathcal{L}$ includes contrastive loss $\mathcal{L}_{c}$ and partial cross-entropy loss $\mathcal{L}_{pce}$, the former is defined in Eq.\ \ref{eq:8}, and the latter is defined as:
\begin{equation}
   \mathcal{L}_{pce}=-\sum_{i\in\tilde{S}}S_{i}log(\hat{S}_{i})+(1-S_{i})log(1-\hat{S}_{i}),
\end{equation}
where $\tilde{S}$ represents the labeled area of $S$. $S_i$ is the true class of pixel $i$, and $\hat{S}_i$ is the prediction of pixel $i$. So the final loss $\mathcal{L}$ can be defined as:
\begin{equation}
   \mathcal{L}=\mathcal{L}_{c}+\mathcal{L}_{pce}.
\end{equation} 

Even though we only use point label in training, our loss is generalizable to the other weakly-supervised COD models by contrastive learning.




\begin{table*}[t]
  \begin{center}
  \caption{Quantitative comparison with state-of-the-arts on four benchmarks. “F”, “U”, “S”, “P” denote fully-supervised, unsupervised, scribble-supervised, and point-supervised methods, respectively. The best results are highlighted in \textbf{bold}.
}\label{tab:1}
  \tabcolsep=0.03cm
    {\scriptsize
\begin{tabular}{l|c|cccc|cccc|cccc}
\toprule[0.8pt]
\multirow{2}*{Methods}&\multirow{2}*{Sup.}&\multicolumn{4}{c|}{CAMO (250)}&\multicolumn{4}{c|}{COD10K (2026)} &\multicolumn{4}{c}{NC4K (4121)}\\

&&MAE $\downarrow$&S$_{m} \uparrow$&E$_{m} \uparrow$&F$^{w}_{\beta} \uparrow$&MAE $\downarrow$&S$_{m} \uparrow$&E$_{m} \uparrow$&F$^{w}_{\beta} \uparrow$&MAE $\downarrow$&S$_{m} \uparrow$&E$_{m} \uparrow$&F$^{w}_{\beta} \uparrow$\\
\toprule[0.8pt]
F3Net~\cite{wei2020f3net}&F&0.109&0.711&0.741&0.564&0.051&0.739&0.795&0.544&0.069&0.782&0.825&0.706\\
CSNet~\cite{gao2020highly}&F&0.092&0.771&0.795&0.641&0.047&0.778&0.809&0.569&0.061&0.819&0.845&0.748\\
ITSD~\cite{zhou2020interactive}&F&0.102&0.750&0.779&0.610&0.051&0.767&0.808&0.557&0.064&0.811&0.845&0.729\\

MINet~\cite{pang2020multi}&F&0.090&0.748&0.791&0.637&0.042&0.77&0.832&0.608&-&-&-&-\\

PraNet~\cite{fan2020pranet}&F&0.094&0.769&0.825&0.663&0.045&0.789&0.861&0.629&-&-&-&-\\
UCNet~\cite{zhang2020uc}&F&0.094&0.739&0.787&0.640&0.042&0.776&0.857&0.633&0.055&0.813&0.872&0.777\\
SINet~\cite{fan2020camouflaged}&F&0.092&0.745&0.804&0.644&0.043&0.776&0.864&0.631&0.058&0.808&0.871&0.723\\
MGL-R~\cite{zhai2021mutual}&F&0.088&0.775&0.812&0.673&0.035&0.814&0.851&0.666&0.052&0.833&0.867&0.740\\
PFNet~\cite{mei2021camouflaged}&F&0.085&0.782&0.841&0.695&0.040&0.800&0.877&0.660&0.053&0.829&0.887&0.745\\
UJSC~\cite{li2021uncertainty}&F&0.073&0.800&0.859&0.728&0.035&0.809&0.884&0.684&0.047&0.842&0.898&0.771\\
UGTR~\cite{yang2021uncertainty}&F&0.086&0.784&0.822&0.684&0.036&0.817&0.852&0.666&0.052&0.839&0.874&0.747\\

ZoomNet\cite{pang2022zoom}&F&0.066&0.820&0.892&0.752&0.029&0.838&0.911&0.729&0.043&0.853&0.896&0.784\\

\hline
DUSD~\cite{zhang2018deep}&U&0.166&0.551&0.594&0.308&0.107&0.580&0.646&0.276&-&-&-&-\\
USPS~\cite{nguyen2019deepusps}&U&0.207&0.568&0.641&0.399&0.196&0.519&0.536&0.265&-&-&-&-\\
SAM~\cite{kirillov2023segment}&U&0.132&0.684&0.687&0.606&0.050&0.783&0.798&\textbf{0.701}&0.078&0.767&0.776&0.696\\
SS~\cite{zhang2020weakly}&S&0.118&0.696&0.786&0.562&0.071&0.684&0.770&0.461&-&-&-&-\\
SCSOD~\cite{yu2021structure}&S&0.102&0.713&0.795&0.618&0.055&0.710&0.805&0.546&-&-&-&-\\
CRNet~\cite{he2023weakly}&S&0.092&0.735&0.815&0.641&0.049&0.733&0.832&0.576&0.063&0.775&0.855&0.688\\
\hline
\rowcolor{lightgray!40}Ours&P&\textbf{0.074}&\textbf{0.798}&\textbf{0.872}&\textbf{0.727}&\textbf{0.042}&\textbf{0.784}&\textbf{0.859}&0.650&\textbf{0.051}&\textbf{0.822}&\textbf{0.889}&\textbf{0.748}\\

\toprule[0.8pt]
\end{tabular}
}
\end{center}
\end{table*}

\begin{figure}[t]
\centering
\includegraphics[width=10cm]{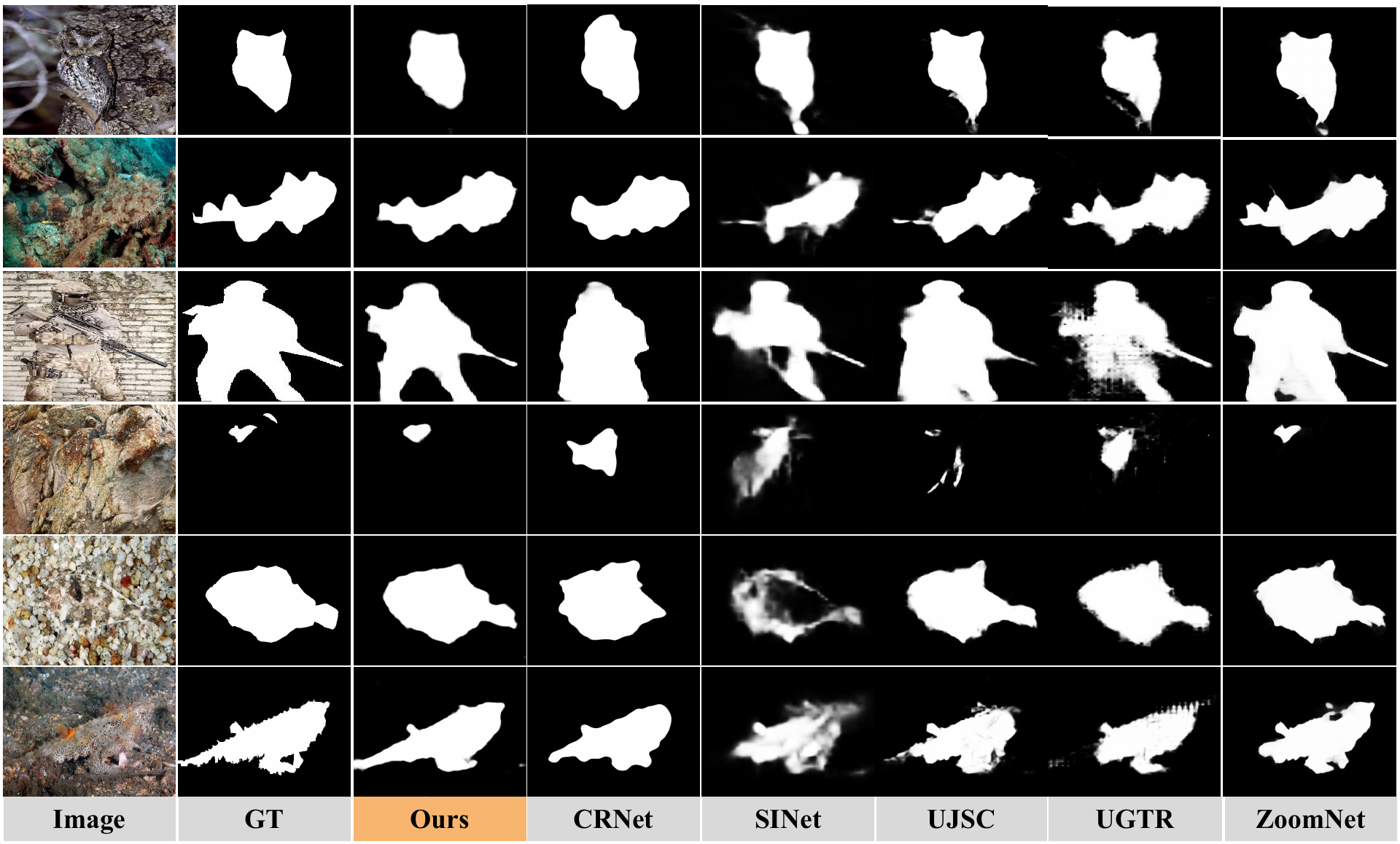}\\
\caption{Visual comparison with representative scribble-supervised and fully-supervised models. }
\label{fig:5}
\end{figure}

\begin{table}[t]
\begin{minipage}{0.5\linewidth}
\tabcolsep=0.15cm
\scriptsize
\centering
\captionsetup{width=.95\textwidth}
\caption{Ablation on point position. }\label{tab:position}
\begin{tabular}{c|cccc}
\toprule[0.8pt]
Type&MAE$\downarrow$&S$_m\uparrow$&E$_m\uparrow$&F$^w_\beta\uparrow$\\
\toprule[0.8pt]
Center&0.079&0.789&0.858&0.716\\
Random&0.076&0.791&0.869&0.722\\
\hline
\rowcolor{lightgray!40}Ours&\textbf{0.074}&\textbf{0.798}&\textbf{0.872}&\textbf{0.727}\\
\toprule[0.8pt]
\end{tabular} 
\end{minipage}\begin{minipage}{0.48\linewidth}  
\tabcolsep=0.1cm
\scriptsize
\centering
\captionsetup{width=.95\textwidth}
\caption{Ablation on point numbers.}\label{tab:number}
\begin{tabular}{c|cccc}
\toprule[0.8pt]
Number of points&MAE$\downarrow$&S$_m\uparrow$&E$_m\uparrow$&F$^w_\beta\uparrow$\\
\toprule[0.8pt]
One-point&0.076&0.791&0.869&0.722\\
Two-points&0.074&0.802&0.881&0.737\\ 
\rowcolor{lightgray!40}Three-points&\textbf{0.074}&\textbf{0.805}&\textbf{0.887}&\textbf{0.741}\\

\toprule[0.8pt]
\end{tabular}
\end{minipage}
\end{table}

\begin{table}[t]
  \begin{center}
  \scriptsize
  \caption{Comparison of parameters and MACs. $^*$ is reset in the same input size.} 
\label{tab:2}
    \tabcolsep=0.15cm
    {\small
\begin{tabular}{c|c|c|cc|cccc}
\toprule[0.8pt]
Methods&Sup.&Input size&Para(M)&MACs(G)&MAE$\downarrow$&S$_{m}$$\uparrow$&E$_{m}$$\uparrow$&F$^{w}_{\beta}$$\uparrow$\\

\toprule[0.8pt]
SINet&F&${352^{2}}$&48.95&19.42&0.092&0.745&0.804&0.644\\
CRNet$^*$&S&${352^{2}}$&32.65&14.27&0.095&0.730&0.804&0.634\\
\hline
\rowcolor{lightgray!40}Ours&P&${352^{2}}$&\textbf{25.44}&\textbf{10.22}&\textbf{0.088}&\textbf{0.761}&\textbf{0.836}&\textbf{0.677}\\
\toprule[0.8pt]
\end{tabular}
}
\end{center}
\end{table}

\begin{table}[t]
\begin{minipage}{0.5\linewidth}
\tabcolsep=0.15cm
\scriptsize
\centering
\captionsetup{width=.95\textwidth}
\caption{Effect of hint area generator with different settings. Point, I, P and SAM denote point label, initial square area, prediction map P and P obtained by SAM as proposal area, respectively.
}\label{tab:3}
\begin{tabular}{l|cccc}
\toprule[0.8pt]
\multicolumn{1}{c|}{Settings}&MAE$\downarrow$&S$_{m}\uparrow$&E$_{m}\uparrow$&F$^{w}_{\beta}\uparrow$\\
\toprule[0.8pt]
w/ Point&\multicolumn{4}{c}{collapse}\\
w/ I&0.092&0.754&0.802&0.655\\
w/ P&0.103&0.726&0.813&0.627\\
w/ SAM&0.108&0.694&0.771&0.603\\
\hline
\rowcolor{lightgray!40}w/ Ours &\textbf{0.085}&\textbf{0.759}&\textbf{0.842}&\textbf{0.677}\\
\toprule[0.8pt]
\end{tabular} 
\end{minipage}\begin{minipage}{0.48\linewidth}
\tabcolsep=0.2cm
\scriptsize
\centering
\captionsetup{width=.95\textwidth}
\caption{Ablation study on contrastive losses. MSE, KL, COS, L1 mean Mean Square Error, Kullback-Leibler divergence, Consine similarity, and L1 loss, respectively.}\label{tab:4}
\begin{tabular}{c|cccc}
\toprule[0.8pt]
Loss&MAE$\downarrow$&S$_m\uparrow$&E$_m\uparrow$&F$^w_\beta\uparrow$\\
\toprule[0.8pt]
w/o&0.085&0.759&0.842&0.677\\
\hline
MSE&0.079&0.791&0.851&0.708\\
KL&0.085&0.769&0.844&0.686\\
COS&0.087&0.766&0.844&0.680\\
\hline
\rowcolor{lightgray!40}L1&\textbf{0.074}&\textbf{0.798}&\textbf{0.872}&\textbf{0.727}\\
\toprule[0.8pt]
\end{tabular}
\end{minipage}

\end{table}

\section{Experiments}

\subsection{Experimental Setup}

\noindent\textbf{Point Dataset.} Our experiments are conducted on
three COD benchmarks, CAMO\ \cite{le2019anabranch}, COD10K\ \cite{fan2020camouflaged}, and NC4K\ \cite{lv2021simultaneously}. In order to evaluate our point-supervised COD method, we relabel 4040 images (3040 from COD10K, 1000 from CAMO) and propose the Point-supervised Dataset (P-COD) for training and the remaining is for testing. In Fig.\ \ref{fig:1}, we show an example of point annotation and compare it with other annotation methods. For each image, we simulate the \textit{hunting process} to label only one point for each camouflaged object, without needing to worry about ambiguous boundaries, making the process of the annotation easy and natural.



\noindent\textbf{Evaluation Metrics.} We adopt four evaluation metrics: mean absolute error (MAE), S-measure (S$_{m}$) \cite{fan2017structure}, E-measure (E$_{m}$) \cite{fan2018enhanced}, weighted F-measure (F$^{w}_{\beta}$) \cite{margolin2014evaluate}.

\noindent\textbf{Implementation Details.} We implement our method with PyTorch and conduct experiments on one GeForce RTX4090 GPU.
We use the stochastic gradient descent optimizer with a momentum of $0.9$, a weight decay of $5e$-$4$, and triangle learning rate schedule with maximum learning rate $1e$-$3$. The batch size is $8$, and the training epoch is $60$. It takes around $7$ hours to train. During training and inference, input images are resized to $512\times512$. 


\subsection{Compare with State-of-the-arts}

\noindent\textbf{Quantitative Comparison.} Being the first point supervised COD method, our proposed approach is primarily benchmarked against scribble-supervised and fully-supervised methods. As demonstrated in Tab.\ \ref{tab:1}, our method achieves substantial improvements, with an average enhancement of $17.6\%$ for MAE, $7.2\%$ for S${m}$, $4.7\%$ for E${m}$, and $11.7\%$ for F$^{w}_{\beta}$ compared to the state-of-the-art weakly-supervised COD method CRNet \cite{he2023weakly}. This highlights our capability to achieve better performance with fewer annotations. Our approach even outperforms fully supervised methods in several metrics across multiple datasets. Specifically, it excels against the majority of fully-supervised methods on CAMO and outpaces nearly half of them on other datasets.


\begin{figure}[t]
\centering
\includegraphics[width=10cm]{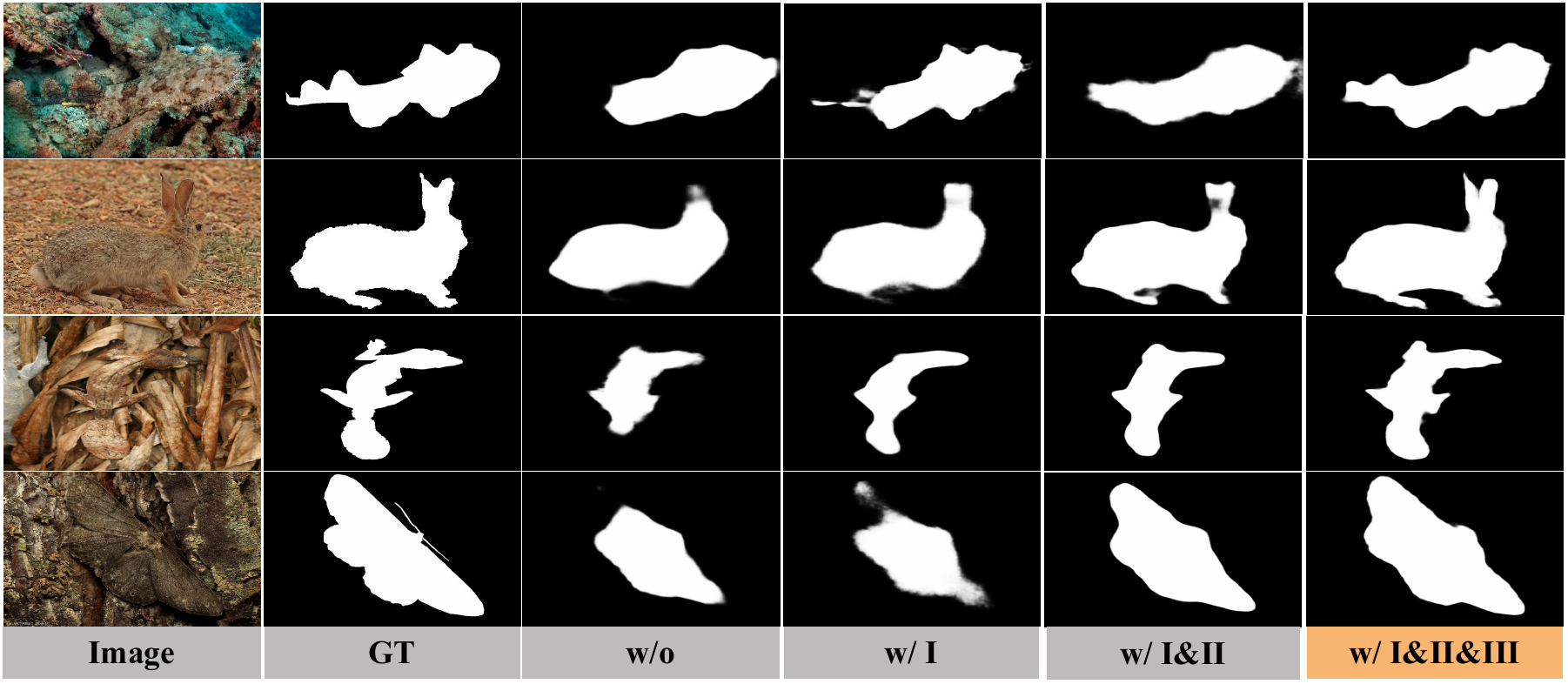}\\
\caption{Visualization of the component ablation study. I, II, III respectively represent hint area generator, representation optimizer, and attention regulator.}
\label{fig:6}
\end{figure}

\noindent\textbf{Qualitative Evaluation.} Our method produces prediction maps characterized by clearer and more complete object regions, along with sharper contours, significantly outperforming state-of-the-art weakly supervised COD method CRNet \cite{he2023weakly}, as shown in Fig. \ref{fig:5}. Our method performs well in various challenging scenarios, including scenarios with tiny objects (row 4), high intrinsic similarities (row 3, 5), indefinable boundaries (row 1), and complex backgrounds (row 2, 6).


\noindent\textbf{Parameter Complexity.} With lower parameter complexity and minimal computational cost, our model (25.44M parameters and 10.22G MACs) consistently outperforms the fully supervised SINet (48.95M parameters and 19.42G MACs) by 4.3\% for MAE, 2.1\% for S${m}$, and 4.0\% for E${m}$ on the challenging CAMO dataset. It also outperforms the scribble supervised CRNet (32.65M parameters and 14.27G MACs) by 7.4\% for MAE, 4.2\% for S${m}$, 4.0\% for E${m}$, and 6.8\% for F$^{w}_{\beta}$, as shown in Tab. \ref{tab:2}. These results underscore the superior lightweight performance of our model.




\begin{table}[t]
  \begin{center}
  \caption{The ablation study for different augmentations in contrastive learning. “S”, “C”, “F”, “T”, “\textbf{R}”, “G”, “J” are Scale, Crop, Flip, Translate, \textbf{Attention Regulator}, Guassblur, Colorjitter.
}\label{tab:6}
   \tabcolsep=0.25cm
    {\scriptsize
\begin{tabular}{ccccccc|cccc}
\toprule[0.8pt]
S&C&F&T&\textbf{R}&G&J&MAE$\downarrow$&S$_{m}\uparrow$&E$_{m}\uparrow$&F$^{w}_{\beta}\uparrow$\\
\toprule[0.8pt]
&&&&&&&0.085&0.759&0.842&0.677\\
\checkmark&&&&&&&0.082&0.779&0.853&0.701\\
\checkmark&\checkmark&&&&&&0.080&0.782&0.860&0.713\\
\checkmark&\checkmark&\checkmark&&&&&0.080&0.771&0.845&0.707\\
\checkmark&\checkmark&\checkmark&\checkmark&&&&0.077&0.781&0.853&0.711\\
\rowcolor{lightgray!40}\checkmark&\checkmark&\checkmark&\checkmark&\checkmark&&&\textbf{0.076}&\textbf{0.790}&\textbf{0.866}&\textbf{0.724}\\
\hline
\checkmark&\checkmark&\checkmark&\checkmark&\checkmark&\checkmark&&0.074&0.794&0.867&\textbf{0.729}\\
\rowcolor{lightgray!40}\checkmark&\checkmark&\checkmark&\checkmark&\checkmark&\checkmark&\checkmark&\textbf{0.074}&\textbf{0.798}&\textbf{0.872}&0.727\\
\toprule[0.8pt]
\end{tabular}
}
\end{center}
\end{table}



\begin{table}[t]
  \begin{center}
  \caption{Effect of different masking setting in Attention regulator.}\label{tab:8}
    \tabcolsep=0.15cm
    {\scriptsize
\begin{tabular}{lc|cccc}
\toprule[0.8pt]
\multicolumn{1}{c}{Basic setting}&Type&MAE$\downarrow$&S$_{m}\uparrow$&E$_{m}\uparrow$&F$^{w}_{\beta}\uparrow$\\
\toprule[0.8pt]
w/ aug (S, C, T, F)&HaS\ \cite{kumar2017hide} &0.078&0.785&0.864&0.711\\
w/ aug (S, C, T, F)&Cutout\ \cite{devries2017improved} &0.076&0.789&0.862&0.721\\
\hline
\rowcolor{lightgray!40}w/ aug (S, C, T, F)&Ours\ \ &\textbf{0.076}&\textbf{0.790}&\textbf{0.866}&\textbf{0.724}\\
\toprule[0.8pt]
\end{tabular}
}
\end{center}
\end{table}



\begin{table}[t]
\begin{minipage}{0.5\linewidth}
\tabcolsep=0.05cm
\scriptsize
\centering
\captionsetup{width=.95\textwidth}
\caption{Effect of key components in optimizer. }\label{tab:7}
\begin{tabular}{cc|cccc}
\toprule[0.8pt]
stop-grad&predictor&MAE$\downarrow$&S$_{m}\uparrow$&E$_{m}\uparrow$&F$^{w}_{\beta}\uparrow$\\
\toprule[0.8pt]
\checkmark&$\times$&\multicolumn{4}{c}{unstable}\\
$\times$&\checkmark&0.081&0.768&0.852&0.706\\
\hline
\rowcolor{lightgray!40}\checkmark&\checkmark&\textbf{0.074}&\textbf{0.798}&\textbf{0.872}&\textbf{0.727}\\

\toprule[0.8pt]
\end{tabular} 
\end{minipage}\begin{minipage}{0.48\linewidth}  
\tabcolsep=0.05cm
\scriptsize
\centering
\captionsetup{width=.95\textwidth}
\caption{Transferability studies on scribble datasets.}\label{tab:10}
\begin{tabular}{c|c|cccc}
\toprule[0.8pt]
Methods&Sup.&MAE$\downarrow$&S$_{m}\uparrow$&E$_m\uparrow$&F$^w_\beta\uparrow$\\
\toprule[0.8pt]
ZoomNet\ \cite{pang2022zoom}&F&0.066&\textbf{0.820}&0.892&0.752\\
CRNet\ \cite{he2023weakly}&S&0.092&0.735&0.815&0.641\\
\hline
\rowcolor{lightgray!40}Ours + scribble&S&\textbf{0.065}&0.816&\textbf{0.894}&\textbf{0.761}\\
\toprule[0.8pt]
\end{tabular}
\end{minipage}
\end{table}



\begin{table}[t]
  \begin{center}
  \caption{The impact of $\tau$, $\alpha$, $w$, and $d$ in the hint area generator.}\label{tab:5}
  \tabcolsep=0.14cm
    {\scriptsize
\begin{tabular}{ccc|ccc|ccc|ccc}
\toprule[0.8pt]
$d$ &MAE$\downarrow$&S$_{m}\uparrow$&$\tau$ &MAE$\downarrow$&S$_{m}\uparrow$&$\alpha$&MAE$\downarrow$&S$_{m}\uparrow$&$w$&MAE$\downarrow$&S$_{m}\uparrow$\\
\toprule[0.8pt]
5&0.085&0.758&100&0.086&0.757&3&0.091&0.739&5&0.089&0.755\\
\textbf{10}&\textbf{0.085}&0.759&\textbf{150}&\textbf{0.085}&\textbf{0.759}&\textbf{4}&\textbf{0.085}&\textbf{0.759}&10&0.087&0.759\\
15&0.086&\textbf{0.760}&200&0.085&0.758&5&0.092&0.736&\textbf{15}&\textbf{0.085}&\textbf{0.759}\\

\toprule[0.8pt]
\end{tabular}
}
\end{center}
\end{table}

\begin{table*}[!t]
  \begin{center}
  \caption{Quantitative comparison with state-of-the-arts on four popular SOD benchmarks. “F”, “S”, “P” denote fully-, scribble-, and point-supervised methods, respectively. The best are in \textbf{bold}. }\label{tab:11}
  \tabcolsep=0.01cm
    {\scriptsize
\begin{tabular}{l|c|ccc|ccc|ccc|ccc}
\toprule[0.8pt]
\multirow{2}*{Methods}&\multirow{2}*{Sup.}&\multicolumn{3}{c|}{ECSSD}&\multicolumn{3}{c|}{DUT-O}&\multicolumn{3}{c|}{HKU-IS} &\multicolumn{3}{c}{DUTS-TE}\\

&&MAE $\downarrow$&S$_{m} \uparrow$&F$^{max}_{\beta} \uparrow$&MAE $\downarrow$&S$_{m} \uparrow$&F$^{max}_{\beta} \uparrow$&MAE $\downarrow$&S$_{m} \uparrow$&F$^{max}_{\beta} \uparrow$&MAE $\downarrow$&S$_{m} \uparrow$&F$^{max}_{\beta} \uparrow$\\
\toprule[0.8pt]





AFNet~\cite{feng2019attentive}&F& 0.042&0.913&0.935 &0.057&0.826 &0.797&0.036&0.905&0.923&0.046&0.867&0.863 \\
BASNet~\cite{qin2019basnet}&F&0.037&0.916&0.943&0.057&0.836&0.805&0.032&0.909&0.928&0.048&0.866&0.859\\
\hline
MFNet~\cite{piao2021mfnet}&S&0.084&0.834&0.879&0.087&0.741&0.706&0.059&0.846&0.876&0.765&0.774&0.770\\
SCSOD~\cite{yu2021structure}&S&0.049&0.881&0.914&0.060&0.811&0.782&0.038&0.882&0.908&0.049&0.853&0.858\\
PSOD~\cite{gao2022weakly}&P&\textbf{0.036}&\textbf{0.913}&\textbf{0.935}&0.064&0.824&0.808&0.033&0.901&0.923&0.045&0.853&0.858\\
\hline
\rowcolor{lightgray!40}Ours&P&0.045&0.895&0.915&\textbf{0.059}&\textbf{0.826}&\textbf{0.824}&\textbf{0.031}&\textbf{0.906}&\textbf{0.926}&\textbf{0.042}&\textbf{0.864}&\textbf{0.861}\\
\toprule[0.8pt]
\end{tabular}
}
\end{center}
\end{table*}





\subsection{Ablation Study} As CAMO dataset is the most challenging one, indicated by the lowest scores in Tab. \ref{tab:1}, all following ablation experiments are performed on it.



\noindent\textbf{Ablation Study on Point Supervision.} 
To reduce annotation costs, we intuitively annotated a point for each camouflaged object by humans.
We conduct experiments on point positions, as shown in Tab. \ref{tab:position}. 
For \textit{Random}, we randomly choose a point from real labels, and repeat three times, taking the average of the results. 
which shows our method far exceeds the center annotation and random annotation under conditions of low annotation cost.  
We attribute it to the model's utilization of the priors implicit in point annotations to simulate human cognitive processes. 
In addition, we conduct experiments for point numbers (\textit{Random}), as shown in Tab. \ref{tab:number}. One point yields satisfactory performance. Increasing the points shows diminishing returns.





\noindent\textbf{Effectiveness of Hint Area Generator.} As shown in Tab.\ \ref{tab:3}, our proposed hint area generator obtains a significant improvement over the direct use of point label, Init square area and Prediction map. Because it has more supervision compared to the Init square area and fewer errors (e.g., background noise) compared to prediction map and prediction map obtained by SAM. Interestingly, the init square area performs much better than prediction map P and P obtained by SAM, and we think that the accuracy of supervision in WSCOD is more crucial than the quantity, since init square area has better accuracy and prediction map has high quantity in area generation but contains a lot of background noise. In addition, compared with prediction map, a relatively complete prediction map can be obtained by using hint area generator, as shown in Fig.\ \ref{fig:6}.




\begin{figure}[t]
\centering
\includegraphics[width=10cm]{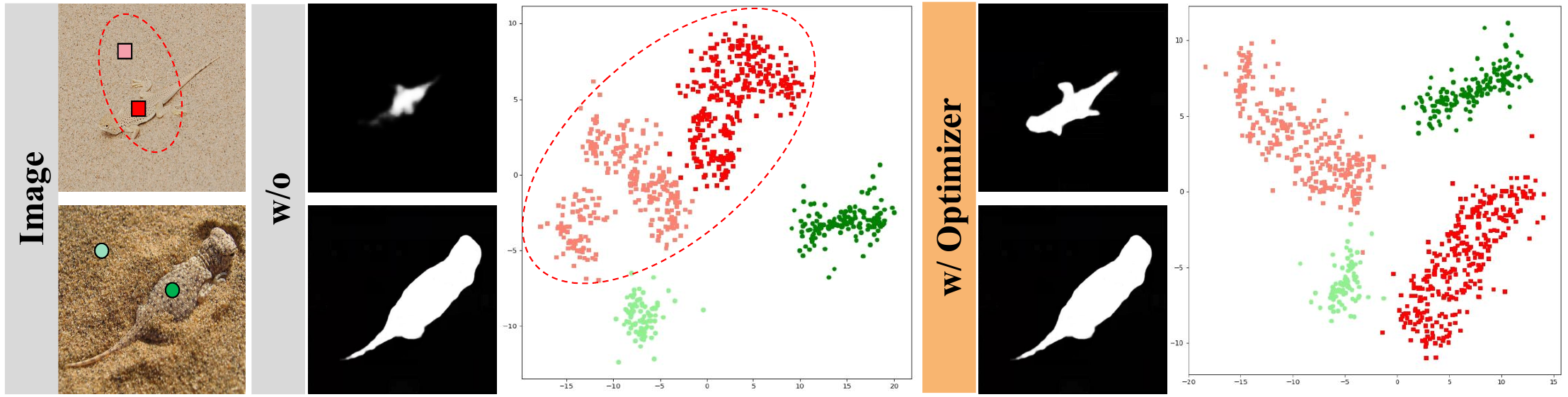}\\
\caption{Visualization of the feature space. Entangled features from foregrounds and backgrounds could be well separated by representation optimizer (visualized by t-SNE). }
\label{fig:7}
\end{figure}

\begin{figure}[t]
\centering
\includegraphics[width=10cm]{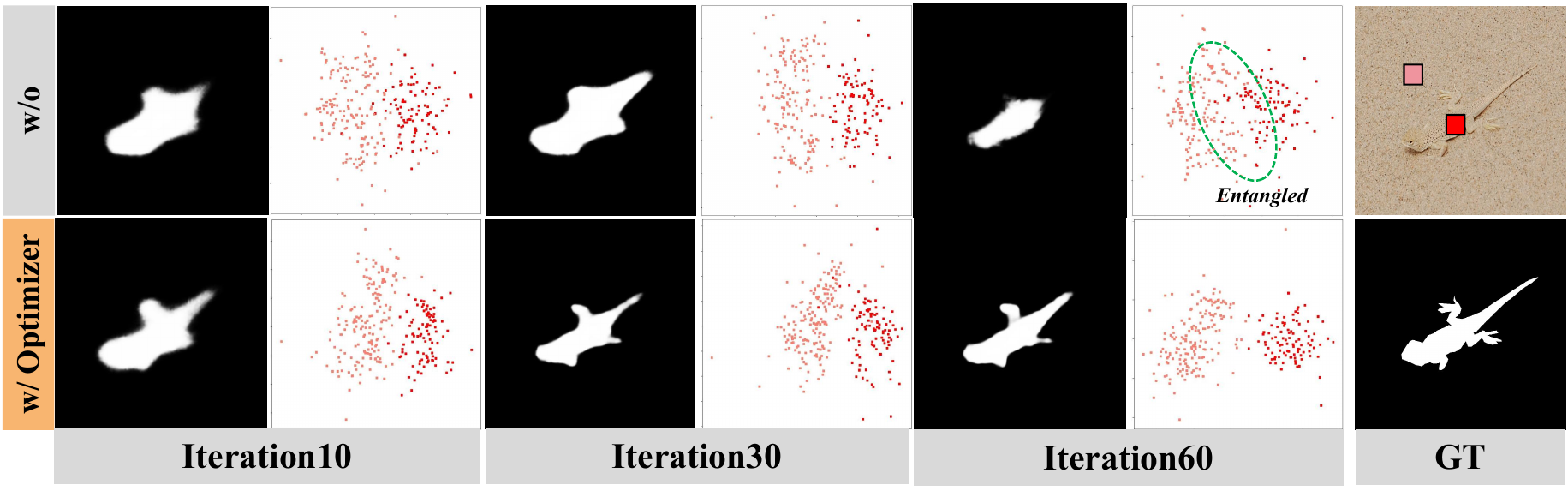}\\
\caption{Visualization of results in different iterations. As training proceeds, the entangled features from the foreground and background can be progressively separated by the representation optimizer (visualization by t-SNE).}
\label{fig:8}
\end{figure}

\noindent\textbf{Effectiveness of Attention Regulator.} We propose attention regulator and apply it with UCL. As shown in Tab. \ref{tab:6}, the results (\textbf{Bold}) are significantly improved after using attention regulator. In addition, to further demonstrate the superiority of regulator in weakly-supervised COD, we compare it with the previous mask approaches HaS \cite{kumar2017hide} and Cutout \cite{devries2017improved}, results are shown in Tab. \ref{tab:8}.

\noindent\textbf{Effectiveness of Representation Optimizer.} We conduct ablation experiments for representation optimizer. Firstly, we test different types of contrastive loss and find that L1 loss performs best, with a significant improvement over the baseline, as shown in Tab.\ \ref{tab:4}. In addition, we conduct exhaustive experiments for data augmentation, which is important for UCL in representation optimizer. Specifically, the augmentation is divided into three aspects, color-texture related (ColorJitter, GuassBlur), position related (Flip, Translate), and size related (Crop, Scale). It shows that color-texture and size augmentation can improve performance, but position augmentation yields unsatisfactory results, as shown in Tab.\ \ref{tab:6}. We employ UCL exclusively with positive samples and test the key components in ablation study, as shown in Tab.\ \ref{tab:7}. Fig.\ \ref{fig:6} intuitively illustrates that the representation optimizer enhances the precision of prediction maps. Fig.\ \ref{fig:7} and \ref{fig:8} show that our model is able to continuously optimize representation through representation optimizer and separate entangled objects from background, making the model eventually learn robust representations.

\noindent\textbf{Impact of the Hyperparameter \textbf{$\tau, \alpha, w, d$.}} Four hyperparameters exist in hint area generator: $\tau$, $\alpha$, $w$, $d$. We test diverse values for these parameters, as shown in Tab.\ \ref{tab:5}. So we set $\tau=150$, $\alpha=4$, $w=15$ and $d=10$. It is worth noting that the values of $\tau$, $w$, and $d$ do not have much impact on the results with different settings. In addition, the $\tau$ can be replaced with unsupervised k-means clustering method (in S.M.). Our model is robust to parameters. It is because the hint area generator is designed to approximate the size of the camouflaged object through the prediction map. Whether the contour is fine and details are perfect, in the prediction map, has little influence on the estimation of the object size. The accurate estimation and localization of the camouflaged object are explored by the attention regulator and the representation optimizer modules. 
\subsection{Transferability Studies} 



\noindent\textbf{Extension to Scribble.} We apply our method to the scribble dataset, and the results are shown in Tab.\ \ref{tab:10} (More results in supplementary materials) reveal that our model, utilizing scribble labeling, achieves competitive results compared to the state-of-the-art fully supervised method ZoomNet\ \cite{pang2022zoom}. It showcases the generalization ability of our method.

\noindent\textbf{Extension to SOD task.} Our method excels not only in COD but also demonstrates competitive performance in SOD, as shown in Tab.\ \ref{tab:11}. Considering the difference between these two tasks, we may wonder \emph{why our method consistently performs well on such two seemingly different tasks}. We attribute this to the generality and rationality of the designed structure. In fact, point-supervised SOD and COD have a clear commonality, i.e., the demands of sufficient and accurate supervision information. Through an effective point label expansion strategy, our model can provide more supervision information than a single point and more accurate supervision information than pseudo-labels, aiding in the precise segmentation of objects. Additionally, the proposed UCL optimizes the representation of weakly-supervised models, reducing segmentation instability. Despite the saliency of objects in WSSOD, it can also benefit from UCL. All these components are built on the common demands of the two tasks, providing a solid foundation for performance.




\section{Conclusion} In this paper, we proposed an effective solution for COD by utilizing point-based supervision. The introduction of the P-COD dataset, based on human cognitive processes, streamlines the annotation process. To overcome the challenge of single-point supervision, we designed a hint area generator bolstering supervision. Representation optimizer optimized model representation by UCL. The attention regulator promoted a holistic object perception, ensuring a comprehensive understanding of camouflaged objects without fixating on distinct parts. Experimental results show that our method outperforms existing weakly-supervised methods largely and even surpasses many fully-supervised ones. 

\noindent{}\textbf{Acknowledgement}. This work is supported by the National Natural Science Foundation of China(NSFC) under Grant 62306239 and the Young Talent Fund of Association for Science and Technology in Shaanxi, China.

\bibliographystyle{splncs04}
\bibliography{main}

\begin{thebibliography}{10}
\providecommand{\url}[1]{\texttt{#1}}
\providecommand{\urlprefix}{URL }
\providecommand{\doi}[1]{https://doi.org/#1}

\bibitem{bearman2016s}
Bearman, A., Russakovsky, O., Ferrari, V., Fei-Fei, L.: What’s the point: Semantic segmentation with point supervision. In: European conference on computer vision. pp. 549--565. Springer (2016)

\bibitem{benenson2019large}
Benenson, R., Popov, S., Ferrari, V.: Large-scale interactive object segmentation with human annotators. In: Proceedings of the IEEE/CVF conference on computer vision and pattern recognition. pp. 11700--11709 (2019)

\bibitem{chen2021exploring}
Chen, X., He, K.: Exploring simple siamese representation learning. In: Proceedings of the IEEE/CVF conference on computer vision and pattern recognition. pp. 15750--15758 (2021)

\bibitem{cheng2022pointly}
Cheng, B., Parkhi, O., Kirillov, A.: Pointly-supervised instance segmentation. In: Proceedings of the IEEE/CVF Conference on Computer Vision and Pattern Recognition. pp. 2617--2626 (2022)

\bibitem{devries2017improved}
DeVries, T., Taylor, G.W.: Improved regularization of convolutional neural networks with cutout. arXiv preprint arXiv:1708.04552  (2017)

\bibitem{fan2017structure}
Fan, D.P., Cheng, M.M., Liu, Y., Li, T., Borji, A.: Structure-measure: A new way to evaluate foreground maps. In: Proceedings of the IEEE international conference on computer vision. pp. 4548--4557 (2017)

\bibitem{fan2018enhanced}
Fan, D.P., Gong, C., Cao, Y., Ren, B., Cheng, M.M., Borji, A.: Enhanced-alignment measure for binary foreground map evaluation. arXiv preprint arXiv:1805.10421  (2018)

\bibitem{fan2021concealed}
Fan, D.P., Ji, G.P., Cheng, M.M., Shao, L.: Concealed object detection. IEEE transactions on pattern analysis and machine intelligence pp. 6024--6042 (2021)

\bibitem{fan2020camouflaged}
Fan, D.P., Ji, G.P., Sun, G., Cheng, M.M., Shen, J., Shao, L.: Camouflaged object detection. In: Proceedings of the IEEE/CVF conference on computer vision and pattern recognition. pp. 2777--2787 (2020)

\bibitem{fan2023advances}
Fan, D.P., Ji, G.P., Xu, P., Cheng, M.M., Sakaridis, C., Van~Gool, L.: Advances in deep concealed scene understanding. Visual Intelligence  \textbf{1}(1), ~16 (2023)

\bibitem{fan2020pranet}
Fan, D.P., Ji, G.P., Zhou, T., Chen, G., Fu, H., Shen, J., Shao, L.: Pranet: Parallel reverse attention network for polyp segmentation. In: International conference on medical image computing and computer-assisted intervention. pp. 263--273. Springer (2020)

\bibitem{fan2020inf}
Fan, D.P., Zhou, T., Ji, G.P., Zhou, Y., Chen, G., Fu, H., Shen, J., Shao, L.: Inf-net: Automatic covid-19 lung infection segmentation from ct images. IEEE transactions on medical imaging pp. 2626--2637 (2020)

\bibitem{feng2019attentive}
Feng, M., Lu, H., Ding, E.: Attentive feedback network for boundary-aware salient object detection. In: Proceedings of the IEEE/CVF conference on computer vision and pattern recognition. pp. 1623--1632 (2019)

\bibitem{perez2012early}
P{\'e}rez-de~la Fuente, R., Delcl{\`o}s, X., Pe{\~n}alver, E., Speranza, M., Wierzchos, J., Ascaso, C., Engel, M.S.: Early evolution and ecology of camouflage in insects. Proceedings of the National Academy of Sciences pp. 21414--21419 (2012)

\bibitem{gao2020highly}
Gao, S.H., Tan, Y.Q., Cheng, M.M., Lu, C., Chen, Y., Yan, S.: Highly efficient salient object detection with 100k parameters. In: European Conference on Computer Vision. pp. 702--721. Springer (2020)

\bibitem{gao2022weakly}
Gao, S., Zhang, W., Wang, Y., Guo, Q., Zhang, C., He, Y., Zhang, W.: Weakly-supervised salient object detection using point supervision. In: Proceedings of the AAAI Conference on Artificial Intelligence. pp. 670--678 (2022)

\bibitem{grill2020bootstrap}
Grill, J.B., Strub, F., Altch{\'e}, F., Tallec, C., Richemond, P., Buchatskaya, E., Doersch, C., Avila~Pires, B., Guo, Z., Gheshlaghi~Azar, M., et~al.: Bootstrap your own latent-a new approach to self-supervised learning. Advances in neural information processing systems  \textbf{33},  21271--21284 (2020)

\bibitem{he2020momentum}
He, K., Fan, H., Wu, Y., Xie, S., Girshick, R.: Momentum contrast for unsupervised visual representation learning. In: Proceedings of the IEEE/CVF conference on computer vision and pattern recognition. pp. 9729--9738 (2020)

\bibitem{he2023weakly}
He, R., Dong, Q., Lin, J., Lau, R.W.: Weakly-supervised camouflaged object detection with scribble annotations. In: Proceedings of the AAAI Conference on Artificial Intelligence. pp. 781--789 (2023)

\bibitem{kirillov2023segment}
Kirillov, A., Mintun, E., Ravi, N., Mao, H., Rolland, C., Gustafson, L., Xiao, T., Whitehead, S., Berg, A.C., Lo, W.Y., et~al.: Segment anything. arXiv preprint arXiv:2304.02643  (2023)

\bibitem{kumar2017hide}
Kumar~Singh, K., Jae~Lee, Y.: Hide-and-seek: Forcing a network to be meticulous for weakly-supervised object and action localization. In: Proceedings of the IEEE International Conference on Computer Vision. pp. 3524--3533 (2017)

\bibitem{le2019anabranch}
Le, T.N., Nguyen, T.V., Nie, Z., Tran, M.T., Sugimoto, A.: Anabranch network for camouflaged object segmentation. Computer vision and image understanding  \textbf{184},  45--56 (2019)

\bibitem{li2021uncertainty}
Li, A., Zhang, J., Lv, Y., Liu, B., Zhang, T., Dai, Y.: Uncertainty-aware joint salient object and camouflaged object detection. In: Proceedings of the IEEE/CVF Conference on Computer Vision and Pattern Recognition. pp. 10071--10081 (2021)

\bibitem{liew2017regional}
Liew, J., Wei, Y., Xiong, W., Ong, S.H., Feng, J.: Regional interactive image segmentation networks. In: 2017 IEEE international conference on computer vision (ICCV). pp. 2746--2754. IEEE (2017)

\bibitem{lv2021simultaneously}
Lv, Y., Zhang, J., Dai, Y., Li, A., Liu, B., Barnes, N., Fan, D.P.: Simultaneously localize, segment and rank the camouflaged objects. In: Proceedings of the IEEE/CVF Conference on Computer Vision and Pattern Recognition. pp. 11591--11601 (2021)

\bibitem{margolin2014evaluate}
Margolin, R., Zelnik-Manor, L., Tal, A.: How to evaluate foreground maps? In: Proceedings of the IEEE conference on computer vision and pattern recognition. pp. 248--255 (2014)

\bibitem{mei2021camouflaged}
Mei, H., Ji, G.P., Wei, Z., Yang, X., Wei, X., Fan, D.P.: Camouflaged object segmentation with distraction mining. In: Proceedings of the IEEE/CVF Conference on Computer Vision and Pattern Recognition. pp. 8772--8781 (2021)

\bibitem{nguyen2019deepusps}
Nguyen, T., Dax, M., Mummadi, C.K., Ngo, N., Nguyen, T.H.P., Lou, Z., Brox, T.: Deepusps: Deep robust unsupervised saliency prediction via self-supervision. Advances in Neural Information Processing Systems  \textbf{32} (2019)

\bibitem{pang2022zoom}
Pang, Y., Zhao, X., Xiang, T.Z., Zhang, L., Lu, H.: Zoom in and out: A mixed-scale triplet network for camouflaged object detection. In: Proceedings of the IEEE/CVF Conference on computer vision and pattern recognition. pp. 2160--2170 (2022)

\bibitem{pang2020multi}
Pang, Y., Zhao, X., Zhang, L., Lu, H.: Multi-scale interactive network for salient object detection. In: Proceedings of the IEEE/CVF conference on computer vision and pattern recognition. pp. 9413--9422 (2020)

\bibitem{piao2021mfnet}
Piao, Y., Wang, J., Zhang, M., Lu, H.: Mfnet: Multi-filter directive network for weakly supervised salient object detection. In: Proceedings of the IEEE/CVF International Conference on Computer Vision. pp. 4136--4145 (2021)

\bibitem{qian2019weakly}
Qian, R., Wei, Y., Shi, H., Li, J., Liu, J., Huang, T.: Weakly supervised scene parsing with point-based distance metric learning. In: Proceedings of the AAAI Conference on Artificial Intelligence. pp. 8843--8850 (2019)

\bibitem{qin2019basnet}
Qin, X., Zhang, Z., Huang, C., Gao, C., Dehghan, M., Jagersand, M.: Basnet: Boundary-aware salient object detection. In: Proceedings of the IEEE/CVF conference on computer vision and pattern recognition. pp. 7479--7489 (2019)

\bibitem{shin2021all}
Shin, G., Xie, W., Albanie, S.: All you need are a few pixels: semantic segmentation with pixelpick. In: Proceedings of the IEEE/CVF International Conference on Computer Vision. pp. 1687--1697 (2021)

\bibitem{wang2023temporal}
Wang, B., Zhao, Y., Yang, L., Long, T., Li, X.: Temporal action localization in the deep learning era: A survey. IEEE Transactions on Pattern Analysis and Machine Intelligence  (2023)

\bibitem{wang2021pyramid}
Wang, W., Xie, E., Li, X., Fan, D.P., Song, K., Liang, D., Lu, T., Luo, P., Shao, L.: Pyramid vision transformer: A versatile backbone for dense prediction without convolutions. In: Proceedings of the IEEE/CVF international conference on computer vision. pp. 568--578 (2021)

\bibitem{wei2021shallow}
Wei, J., Wang, Q., Li, Z., Wang, S., Zhou, S.K., Cui, S.: Shallow feature matters for weakly supervised object localization. In: Proceedings of the IEEE/CVF Conference on Computer Vision and Pattern Recognition. pp. 5993--6001 (2021)

\bibitem{wei2020f3net}
Wei, J., Wang, S., Huang, Q.: F$^3$net: fusion, feedback and focus for salient object detection. In: Proceedings of the AAAI conference on artificial intelligence. pp. 12321--12328 (2020)

\bibitem{yang2021uncertainty}
Yang, F., Zhai, Q., Li, X., Huang, R., Luo, A., Cheng, H., Fan, D.P.: Uncertainty-guided transformer reasoning for camouflaged object detection. In: Proceedings of the IEEE/CVF International Conference on Computer Vision. pp. 4146--4155 (2021)

\bibitem{yu2021structure}
Yu, S., Zhang, B., Xiao, J., Lim, E.G.: Structure-consistent weakly supervised salient object detection with local saliency coherence. In: Proceedings of the AAAI conference on artificial intelligence. pp. 3234--3242 (2021)

\bibitem{zhai2021mutual}
Zhai, Q., Li, X., Yang, F., Chen, C., Cheng, H., Fan, D.P.: Mutual graph learning for camouflaged object detection. In: Proceedings of the IEEE/CVF Conference on Computer Vision and Pattern Recognition. pp. 12997--13007 (2021)

\bibitem{zhang2020uc}
Zhang, J., Fan, D.P., Dai, Y., Anwar, S., Saleh, F.S., Zhang, T., Barnes, N.: Uc-net: Uncertainty inspired rgb-d saliency detection via conditional variational autoencoders. In: Proceedings of the IEEE/CVF conference on computer vision and pattern recognition. pp. 8582--8591 (2020)

\bibitem{zhang2020weakly}
Zhang, J., Yu, X., Li, A., Song, P., Liu, B., Dai, Y.: Weakly-supervised salient object detection via scribble annotations. In: Proceedings of the IEEE/CVF conference on computer vision and pattern recognition. pp. 12546--12555 (2020)

\bibitem{zhang2018deep}
Zhang, J., Zhang, T., Dai, Y., Harandi, M., Hartley, R.: Deep unsupervised saliency detection: A multiple noisy labeling perspective. In: Proceedings of the IEEE conference on computer vision and pattern recognition. pp. 9029--9038 (2018)

\bibitem{zhou2020interactive}
Zhou, H., Xie, X., Lai, J.H., Chen, Z., Yang, L.: Interactive two-stream decoder for accurate and fast saliency detection. In: Proceedings of the IEEE/CVF conference on computer vision and pattern recognition. pp. 9141--9150 (2020)

\end{thebibliography}
\end{document}